\documentclass[11pt,a4paper]{article}
\usepackage[hyperref]{acl2021}
\usepackage{times}
\usepackage{latexsym}

\usepackage{graphicx} 

\usepackage{microtype}
\usepackage{CJKutf8}
\aclfinalcopy 
\def\aclpaperid{46} 

\title{Using Perturbed Length-aware Positional Encoding \\for Non-autoregressive Neural Machine Translation}

\author{Yui Oka\thanks{Currently with NTT Communication Science Laboratories. This work was completed when the first author was a graduate student.}, Katsuhito Sudoh, \and Satoshi Nakamura \\
         Nara Institute of Science and Technology \\
         {\tt yui.oka.vf@hco.ntt.co.jp, \{sudoh, s-nakamura\}@is.naist.jp}}

\date{}

\begin{document}
\maketitle
\begin{abstract}
Non-autoregressive neural machine translation (NAT) usually employs sequence-level knowledge distillation using autoregressive neural machine translation (AT) as its teacher model.
However, a NAT model often outputs shorter sentences than an AT model.
In this work, we propose sequence-level knowledge distillation (SKD) using perturbed length-aware positional encoding and apply it to a student model, the Levenshtein Transformer.
Our method outperformed a standard Levenshtein Transformer by 2.5 points in bilingual evaluation understudy (BLEU) at maximum in a WMT14 German to English translation.
The NAT model output longer sentences than the baseline NAT models.
\end{abstract}

\section{Introduction}
A neural machine translation model (NMT) often outputs sentences shorter than their references.
Various approaches have been proposed to combat such short output problems.
For example, \newcite{DBLP:conf/aaai/ZhaoZZHW19} proposed a method to reduce the entropy of high entropy words.
They observed that source language words with high entropy were not translated in short output problems, such as in under-translation situations.
By reducing the entropy of high-entropy source words and training for correct translations, they helped solve the under-translation problem.

Another approach constrained the output length by directly using length-aware positional encoding (PE) \cite{oka-etal-2020-incorporating}.
It tackled the under-translation problem with output length constraints using length-aware PE and length prediction.
The length constraints are given by the sentence lengths in training and the predicted output lengths in inference.
They proposed a method that adds perturbation to length-aware PEs to relax strict length constraints
and improved the bilingual evaluation understudy (BLEU) and output lengths against a baseline using a standard sinusoidal PE.

Short translation outputs often occur not only with autoregressive (AT) models but also in non-autoregressive (NAT) models with NAT usually suffering more seriously than AT.
Table \ref{example} shows an example where NAT suffered by dropping the verb phrase “were used” and output a length-ratio (LR) smaller sentence.
Recent studies on NAT usually use sequence-level knowledge distillation (SKD) \cite{kim-rush-2016-sequence} to mitigate this problem.
In SKD, an autoregressive Transformer is used as a teacher model to transfer knowledge to a weak student model; a student NAT model is trained to mimic the teacher model’s outputs.
\newcite{Zhou2020Understanding} reported that the accuracy of this teacher model affects the translation accuracy of NAT models.

\begin{center}
\begin{table}[t]
\small
\begin{CJK}{UTF8}{min}
\begin{center}
\begin{tabular}{l|l}
\hline \hline 
Src & For the coils, here {\bf were used} two coils \\
& with 6 mm of inner diameter and 800 \\
& of coil number by inversely connecting.\\ \hline
Ref & ▁ コイル としては , 内径 6 mm で 800 ターン\\
& の 2 基 の コイル を 逆 接続 して \underline{用いた 。}\\ 
AT & ▁ コイル は 内 径 6 mm , コイル 枚 数 800 本\\
& の 2 コイル を 逆 接続 して \underline{使用した 。}\\
NAT & ▁ コイル は 内 径 6 mm , コイル 数 800 の \\
& 二つの コイル を 逆 接続 した 。\\
\hline \hline
\end{tabular}
\end{center}
\caption{\label{example} Example of excessively short output by NAT; Standard AT outputs '\underline{使用した。},' which means '{\bf were used},' but NAT cannot output this.}
\end{CJK}
\end{table}
\end{center}

This work focuses on the output length by SKD-trained NAT.
We propose using perturbation in length-aware PE for both the SKD and NAT models.
In SKD, the AT model as a teacher model can be constrained by the given reference length, a step that is expected to improve the accuracy of the AT results used for SKD.
We can also apply perturbed length-aware PE to the Levenshtein Transformer as a student model to encourage longer outputs than the baseline NAT model.
Our experimental results improved translation accuracy over the baseline NAT with a standard SKD in English to Japanese and German to English translations.

\section{Related Work} 


\subsection{Perturbation into Length-aware Positional Encoding}
\label{secldpe}
\citet{oka-etal-2020-incorporating} incorporated random perturbation into the length constraints for length-difference PE (LDPE) \cite{takase-okazaki-2019-positional}.
The perturbation is given as a random integer from a uniform distribution within a certain range in the training time.
The perturbed LDPE (perLDPE) is added as follows:
\begin{equation}
  perLDPE_{(pos,len,2i)} = sin\left( \frac{len-pos+per}{10000^\frac{2i}{d}} \right)
\end{equation}
\begin{equation}
  perLDPE_{(pos,len,2i+1)} = cos\left( \frac{len-pos+per}{10000^\frac{2i}{d}} \right),
\end{equation}
where $pos$ is the absolute position in the sequence, $2i$ and $2i+1$ represent even and odd dimensions in the PE vector, $d$ is the  embedding dimension, $len$ is the given output sequence length, and $per$ is the given random perturbation.
The perLDPE is only used in the decoder.
\citet{oka-etal-2020-incorporating} used the predicted length with Bidirectional Encoder Representations from Transformers (BERT) \cite{DBLP:journals/corr/abs-1810-04805} as generation length constraints in English to Japanese translation.
The translation accuracy improved significantly using oracle length constraints.

\subsection{Knowledge Distillation in Non-autoregressive Translation}
Knowledge distillation \cite{44873} is a method that uses the distilled knowledge learned by a stronger teacher model in the learning of a weaker student model.
SKD gives a student model the output of a teacher model as knowledge. SKD propagates a wide range of knowledge by the teacher model to the student model and trains it to mimic its knowledge \cite{kim-rush-2016-sequence}.
NAT models rely on distilled data from SKD using AT models as teacher models.

\section{Proposed Method}
Motivated by Section \ref{secldpe}, we propose two methods using length control for translation with the NAT model.
(1) We apply length control to the AT Transformer as the teacher model in SKD, and
(2) Incorporate length-aware PE into the NAT model.


\subsection{SKD using Perturbed Length-aware Positional Encoding}
\label{SKD}
A standard autoregressive Transformer is usually used for a teacher model in SKD for NAT.
Since a standard Transformer often generates short sentences, this problem generally occurs even with distilled data.
We incorporated perturbed length-aware PE into a teacher Transformer model 
to improve the quality of its outputs for better knowledge distillation because we can use ideal length constraints from the target language sentences during training.
The perturbation applies only to training. The perturbed and non-perturbed LDPE are only used in the decoder.

\subsection{NAT using Perturbed Length-aware Positional Encoding}
\label{student}
We employed the Levenshtein Transformer \cite{NIPS2019_9297} as the NAT model.
It has three decoders to insert placeholders and a word into each placeholder token and to delete unnecessary tokens.
The encoder and the decoders have position embeddings.
Although most NAT models output fixed-length sentences, the Levenshtein Transformer iteratively changes the output length by deletion and insertion.
As demonstrated by the empirical study in Section \ref{Result}, the Levenshtein Transformer \cite{NIPS2019_9297} often outputs a model that is shorter than the AT model.
For this problem, we only incorporate perturbed length-aware PE into the placeholder decoder, which is considered length manipulation without sentence content.
This perturbation is used only in the training time, as in the methods mentioned above.
Note that the other two decoders still use position embeddings in the proposed method.

\section{Experiments}
We experimentally evaluated the performance of our proposed method and compared it to the existing methods.

\begin{center}
\begin{table*}[t]
\begin{center}
\begin{tabular}{lc|cc|cc|cc}
\hline \hline 
    & & \multicolumn{2}{c|}{$En\rightarrow Ja$} & \multicolumn{2}{c|}{$En\rightarrow De$} & \multicolumn{2}{c}{$De\rightarrow En$} \\
Model       & emb & BLEU & LR & BLEU & LR & BLEU & LR \\ \hline
Transformer &    & 37.1 & 0.948 & 31.0 & 0.960 & 33.0 & 0.908 \\ \hline
\multicolumn{8}{c}{SKD model: standard Transformer w/ sinusoidal PE (baseline)} \\ \hline
MaskT          & shared & 31.0 & 0.928 & 25.9 & 0.975 & 28.8 & 0.880 \\
LevT           & shared & 34.0 & 0.912 & 28.7 & 0.905 & 27.4 & 0.838 \\
LevT + perLDPE & shared & 33.2 & 0.897 & 26.2 & 0.989 & 29.4 & 0.959 \\
LevT + perLDPE & independent & 34.1 & 0.920 & 26.9 & 0.955 & 28.7 & 0.956 \\ \hline
\multicolumn{8}{c}{SKD model: Transformer w/ perLDPE $[-4,4]$ (proposed)} \\ \hline
MaskT          & shared & \bf 31.3 & 0.943 & 25.9 & 0.955 & 28.3 & 0.884 \\
LevT           & shared & \bf 34.3 & 0.900 & 27.4 & 0.919 & 28.0 & 0.839 \\
LevT + perLDPE & shared & 34.0 & 0.918 & 26.3 & 0.928 & 29.5 & 0.951 \\
LevT + perLDPE & independent & 34.2 & 0.922 & 25.5 & 0.966 & \bf 29.9 & 0.941 \\ \hline \hline
\end{tabular}
\end{center}
\caption{\label{bleu1} Bilingual evaluation understudy (BLEU) and length-ratio (LR) results with different sequence-level knowledge distillation (SKD) and different student models; BLEU values in {\bf bold} most outperformed the baseline. $LevT$ stands for Levenshtein Transformer and $MaskT$ stands for conditional masked language model (CMLM) with Mask-Predict model.}
\end{table*}
\end{center}

\subsection{Settings}
\label{setting}
We used three translation tasks for the experiments: English to Japanese (En-Ja) using the Asian Scientific Paper Excerpt Corpus (ASPEC) \cite{NAKAZAWA16.621}, and English to German (En-De) and German to English (De-En) using WMT14 \cite{bojar-EtAl:2014:W14-33}.
From the ASPEC dataset, we used the first 1 million sentence pairs of the training set with 1,784 and 1,812 sentence pairs for the development and test sets.
The WMT14 dataset consisted of 4.4 million sentence pairs for training a pre-processed one distributed by the Stanford Natural Language Processing (NLP) group\footnote{\url{https://nlp.stanford.edu/projects/nmt/}}. 
We chose newstest 2013 (3,000 sentence pairs) and newstest 2014 (2,737 sentences) for the development and test sets.
All sentences were tokenized into subwords using a SentencePiece model \cite{kudo-richardson-2018-sentencepiece} with a shared subword vocabulary of 16,000 entries in ASPEC and 30,000 entries in WMT14.
The length-aware PE used subword-based lengths in all experiments.

All models were implemented based on fairseq \cite{ott2019fairseq}.
The hyperparameter settings were from fairseq NAT examples \footnote{\url{https://github.com/pytorch/fairseq/blob/master/examples/nonautoregressive_translation/README.md}} for both AT and NAT, except for the number of training epochs (50) and the batch size (18,000).

\begin{center}
\begin{table*}[t]
\begin{center}
\begin{tabular}{lc|cc|cc|cc}
\hline \hline 
\multicolumn{8}{c}{Length Constraints in Student Model: {\bf Reference Length} (\emph{correct})} \\ \hline \hline
& & \multicolumn{2}{c|}{$En\rightarrow Ja$} & \multicolumn{2}{c|}{$En\rightarrow De$} & \multicolumn{2}{c}{$De\rightarrow En$} \\
Model       & emb & BLEU & LR & BLEU & LR & BLEU & LR \\ \hline
Transformer &    & 37.1 & 0.948 & 31.0 & 0.960 & 33.0 & 0.908 \\ \hline
\multicolumn{8}{c}{SKD model: Standard Transformer w/ sinusoidal PE (baseline)} \\ \hline
LevT (baseline)& shared & 34.0 & 0.912 & 28.7 & 0.905 & 27.4 & 0.838 \\
LevT + perLDPE & shared & \bf 34.2 & 0.951 & \bf 30.0 & 0.997 & \bf 32.6 & 0.954 \\
LevT + perLDPE & independent & \bf 34.6 & 0.975 & \bf 31.0 & 0.962 & \bf 32.1 & 0.950 \\ \hline
\multicolumn{8}{c}{SKD model: Transformer w/ perLDPE $[-4,4]$ (proposed)} \\ \hline
LevT + perLDPE & shared & \bf 34.5 & 0.988 & \bf 30.0 & 0.934 & \bf 32.7 & 0.946 \\
LevT + perLDPE & independent & \bf 34.3 & 0.989 & \bf 29.1 & 0.970 & \bf 32.6 & 0.934 \\ \hline \hline
\end{tabular}
\end{center}
\caption{\label{bleu-ref} Bilingual evaluation understudy (BLEU) and length-ratio (LR) results with different sequence-level knowledge distillation (SKD) and different student models using {\bf reference length as length constraints}; BLEU values in {\bf bold} outperformed the baseline.}
\end{table*}
\end{center}

\begin{figure}[t]
\centering
\includegraphics[width=6cm]{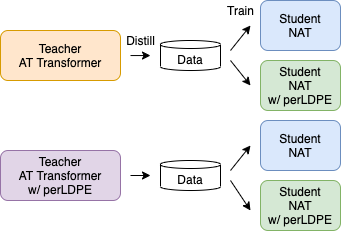}
\caption{Comparison of training process using the baseline and proposed in the SKD and NAT models, respectively.}
\label{trainingprocess}
\end{figure}

\subsection{Models}
Figure \ref{trainingprocess}shows the training process using the baseline and the proposed.
The models are: 
\begin{itemize}
   \setlength{\itemsep}{-5pt}
\item Teacher AT Transformer (AT baseline, $Transformer$)
\item Teacher AT Transformer with LDPE ($Transformer$ $w/ perLDPE$)
\item Student NAT model and SKD using $Transformer$ (NAT baseline, $LevT$, and $MaskT$)
\item Student NAT model and SKD using $Transformer$ $w/perLDPE$
\item Student NAT model with LDPE (perturbation in training, $LevT+perLDPE$), and SKD using $Transformer$ 
\item Student NAT model with LDPE (perturbation in training, $LevT+perLDPE$), and SKD using $Transformer$ $w/ perLDPE$
\end{itemize}

We used a standard Transformer with sinusoidal PE as a teacher AT model baseline in the knowledge distillation, a standard Levenshtein Transformer \cite{NIPS2019_9297}, and a conditional masked language model (CMLM) with Mask-Predict \cite{ghazvininejad-etal-2019-mask} as a student NAT model.
Each of the student models shared embedding parameters.
We applied the proposed SKD described in \ref{SKD} to two different NAT models: LevT and CMLM with Mask-Predict.
We also applied the proposed NAT described in \ref{student} using LevT.

\subsection{Perturbation Range}
\label{persec}
We used perturbation ranges of $[-4,4]$ for the teacher AT model and $[0,2]$ for the student NAT models.
We did not give negative perturbations to the student model to encourage long outputs.
We also compared two different conditions in the embedding parameter matrices in the student NAT models: shared and independent.

\subsection{Length Constraints in Inference}
The proposed NAT model described in Section \ref{student} needs length prediction.
For the length constraints of our student models in inference, we used BERT-based length prediction \cite{oka-etal-2020-incorporating} for En-Ja and a proxy by input length \cite{lakew_iwslt19} for En-De and De-En.

\subsection{Evaluation Metrics}
We used BLEU \cite{papineni-etal-2002-bleu} as our main quality evaluation metric.
All BLEU scores were calculated by sacreBLEU \cite{post-2018-call}.
In the En-Ja translation, the translation results were re-tokenized by MeCab \cite{10019716933} after subword detokenization.
We also investigated the LR of the output and reference sentences in the subword level to evaluate the output length.

\subsection{Results}
\label{Result}
Table \ref{bleu1} shows the BLEU and LR results.
The latter showed that the baseline Levenshtein Transformer with the standard SKD resulted in shorter sentences than the Transformer.
In the En-Ja and De-En experiments, the proposed method outperformed the baseline Levenshtein Transformer.
However, BLEU decreased when we used shared embeddings in the En-Ja translation, although BLEU and LR improved when we used the independent embeddings.
The De-En experiment results were different from the others; BLEU improved even with shared embeddings.
The baseline Levenshtein Transformer resulted in a smaller LR in De-En than in the other tasks and many under-translations occurred.
However, the proposed method did not outperform the baseline in the En-De translation.

On the other hand, the Mask-Predict model with the proposed SKD outperformed the baseline in the En-Ja translation; however, it was ineffective in the other tasks.
Further investigation will be conducted in our future research.

\section{Analysis}

\paragraph{Oracle length constraints}
We investigated the translation accuracy of the student model using oracle length constraints in the generation.
Table \ref{bleu-ref} shows the BLEU and LR results.
All of the proposed models outperformed the baseline Levenshtein Transformer.
In the En-De experiment, the proposed method with a base SKD and independent embedding had the same BLEU as the base Transformer.
The proposed method can be improved if it can achieve better length prediction.

\paragraph{Teacher model}
Table \ref{bleu2} shows the BLEU results by the autoregressive Transformer as the SKD’s teacher model.
Similar to the results shown by \citet{oka-etal-2020-incorporating}, BLEU improved significantly due to the help of \emph{correct} length constraints.
According to \newcite{Zhou2020Understanding}, improving the translation accuracy of the teacher model raises the translation accuracy of the student model. 
However, our results were mixed; no such tendency was observed in the En-De translation or with Mask-Predict.

\begin{table}[t]
\begin{center}
\small
\begin{tabular}{lccc}
\hline \hline
Model & \small $En\rightarrow Ja$ & \small $En\rightarrow De$ & \small $De\rightarrow En$ \\ \hline
Transformer & 32.4 & 30.1 & 32.9 \\ 
w/ perLDPE & \bf 32.5 & \bf 31.1 & \bf 34.9 \\ \hline \hline
\end{tabular}
\end{center}
\caption{\label{bleu2} BLEU results in training set using teacher AT models. In \citet{oka-etal-2020-incorporating}, we used the reference length as length constraints.}
\end{table}

\begin{table}[t]
\begin{center}
\small
\begin{tabular}{r|ccc}
\hline \hline
\multicolumn{1}{l}{Model}  & constraints & BLEU & LR \\ \hline
\multicolumn{4}{c}{ASPEC $En\rightarrow Ja$} \\ \hline
\multicolumn{1}{l}{LevT (baseline)} &  & 34.0 & 0.909 \\ \hline
LevT + perLDPE$[0,2]$  & predict & 34.1 & 0.920\\
+ perLDPE$[0,4]$  &  & 33.2 & 0.900 \\
+ perLDPE$[0,6]$  &  & \textbf{34.2} & 0.919 \\
LevT + perLDPE$[0,2]$  & reference & \underline{34.6} & 0.975\\ 
+ perLDPE$[0,4]$  &  & 33.9 & 0.940 \\
+ perLDPE$[0,6]$  &  & \underline{34.5} & 0.957 \\\hline
\multicolumn{4}{c}{WMT14 $En\rightarrow De$}  \\ \hline
\multicolumn{1}{l}{LevT (baseline)} & & 28.7 & 0.976 \\ \hline
LevT + perLDPE$[0,2]$  & source & 26.9 & 0.955\\
+ perLDPE$[0,4]$  &  & 25.1 & 0.955\\
+ perLDPE$[0,6]$  &  & 26.0 & 0.935\\
LevT + perLDPE$[0,2]$  & reference & \underline{31.0} & 0.962\\ 
+ perLDPE$[0,4]$  &  & \underline{28.8} & 0.956\\
+ perLDPE$[0,6]$  &  & \underline{30.0} & 0.938\\
\hline \hline
\end{tabular}
\end{center}
\caption{\label{bleuallper} BLEU and length-ratio (LR) results with {\bf different perturbation range} using different lengths as length constraints; BLEU values in {\bf bold} outperformed the baseline and \underline{underline} outperformed the baseline when using the reference length as length constraints.}
\end{table}

\paragraph{Perturbation range}
We also investigated how the perturbation range affected the translations by NAT models with baseline SKD using a standard Transformer.
We used the perturbation ranges [0,4] and [0,6] in addition to Section \ref{persec} and used only the model with independent embedding.
Table \ref{bleuallper} shows the results in En-Ja and En-De tasks.
In the En-Ja translation, the model with the range of [0,6] and prediction-based length constraints outperformed the baseline.
However, larger perturbation did not significantly outperform the model with [0,2].
On the other hand, when we used oracle length constraints, the BLEU score and LR dropped by larger perturbation ranges.
This was also the case for En-De translation.
Unlike \newcite{oka-etal-2020-incorporating}, the larger perturbation did not affect the NAT model.

\section{Conclusion}
We incorporated perturbed length-aware PE into the SKD and Levenshtein Transformer.
The experimental results showed BLEU improvements in ASPEC En-Ja and WMT De-En translations, but not in WMT En-De due to inaccurate length constraints.\footnote{We discussed this issue in another paper \cite{okanlp} (in press). We reported WMT14 En-De and De-En results using some length constraints including length prediction and AT Transformer with perLDPE. We also discussed other perturbation ranges in this paper.}
We also investigated translation accuracy using the oracle length as length constraints and identified promising results for further improvement due to more accurate output length predictions.
Future work requires more accurate length predictions, which we expect to be effective based on our analyses using oracle length constraints.

\section*{Acknowledgments}
Part of this work was supported by JSPS KAKENHI Grant Number JP17H06101.

\bibliographystyle{acl_natbib}
\bibliography{anthology,acl2021}


\end{document}